\documentclass[11pt, a4paper, logo, twocolumn, copyright]{deepmind}

\usepackage{amsmath}
\usepackage{graphicx}
\usepackage{float}
\usepackage{algorithm}
\usepackage{algpseudocode}
\usepackage{booktabs}
\usepackage{lettrine}
\usepackage{authblk}
\usepackage{natbib}
\usepackage{xcolor}
\usepackage{subcaption}
\usepackage[title,toc,titletoc,page]{appendix}

\newcommand{\colorhref}[3][blue]{\href{#2}{\color{#1}{#3}}}%

\reportnumber{001} 

\usepackage{pgfplots}
\pgfplotsset{compat=1.17}
\usepackage{tikz}
\usepackage{tikzscale}
\usetikzlibrary{matrix}
\usetikzlibrary[pgfplots.groupplots]
\usetikzlibrary{backgrounds}

\newcommand{\cost}{{c}}
\newcommand{\precost}{{l}}
\newcommand{\state}{{x}}
\newcommand{\nextstate}{{y}}
\newcommand{\control}{{u}}
\newcommand{\objective}{{J}}
\newcommand{\residual}{{r}}
\newcommand{\app}{{MJPC}}

\newcommand{\plan}{{\mathbf{\Pi}}}

\title{Predictive Sampling:\\Real-time Behaviour Synthesis with MuJoCo}

\correspondingauthor{thowell@stanford.edu, tassa@deepmind.com}
\author[1,2]{Taylor Howell}
\author[2]{Nimrod Gileadi}
\author[2]{Saran Tunyasuvunakool}
\author[2,3]{Kevin Zakka}
\author[2]{Tom Erez}
\author[2]{Yuval Tassa}
\affil[1]{Stanford University}
\affil[2]{DeepMind}
\affil[3]{University of California Berkeley}
\renewcommand\Affilfont{\itshape\small}

\begin{document}
	
	\maketitle
	We introduce MuJoCo MPC (\app{}), an open-source, interactive application and software framework for real-time predictive control, based on MuJoCo physics. \app{} allows the user to easily author and solve complex robotics tasks, and currently supports three shooting-based planners: derivative-based iLQG and Gradient Descent, and a simple derivative-free method we call Predictive Sampling. Predictive Sampling was designed as an elementary baseline, mostly for its pedagogical value, but turned out to be surprisingly competitive with the more established algorithms. This work does not present algorithmic advances, and instead, prioritises performant algorithms, simple code, and accessibility of model-based methods via intuitive and interactive software. \app{} is available at \colorhref{https://github.com/deepmind/mujoco_mpc}{github.com/deepmind/mujoco\_mpc}, a video summary can be viewed at \colorhref{https://dpmd.ai/mjpc}{dpmd.ai/mjpc}.

	\section{Introduction} \label{introduction}

\lettrine[lines=2]{M}{odel-based} approaches form the foundation of classical control and robotics. Since Kalman's seminal work~\citep{kalman1960general}, the \emph{state} along with its dynamics and observation models has played a central role. 

The classical approach is being challenged by \textit{learning-based} methods, which forgo the explicit description of the state and associated models, letting internal representations emerge from the learning process \citep{lillicrap2015continuous, schulman2017proximal, salimans2017evolution, smith2022walk, rudin2022learning}. The flexibility afforded by learned representations makes these methods powerful and general, but the requirement for large amounts of data and computation makes them slow. In contrast, pure model-based methods, like the ones described below, can synthesise behaviour in real time~\citep{tassa2014control}.

Since both approaches ultimately generate behaviour by optimising an objective, there is reason to believe they can be effectively combined. Indeed, well-known discrete-domain breakthroughs like AlphaGo~\citep{silver2016mastering} are predicated on combining model-based search and learning-based value and policy approximation. We believe the same could happen for robotics and control, and describe our thinking on how this might happen in the Discussion (Section \ref{discussion}). However, before the community can rise to this challenge, a core problem must be overcome: Model-based optimisation is difficult to implement, often depends on elaborate optimisation algorithms, and is generally inaccessible. 

To address this deficit, we present \textit{\app{}}, an open-source interactive application and software framework for predictive control, based on MuJoCo physics~\citep{todorov2012mujoco}, which lets the user easily author and solve complex tasks using predictive control algorithms in real time. The tool offers implementations of standard derivative-based algorithms: iLQG (second-order planner) and Gradient Descent (first-order planner).  Additionally, it introduces \emph{Predictive Sampling}, a simple zero-order, sampling-based algorithm that works surprisingly well and is easy to understand. 

Importantly, the interactive simulation can be slowed down asynchronously, speeding up the planner with respect to simulation time. This means that behaviours can be generated on older, slower machines, leading to a democratisation of predictive control tooling.

\app{} and Predictive Sampling advance our central goal of lowering the barriers to entry for predictive control in robotics research. An important secondary goal is to accelerate \textit{research velocity}. When tweaking a parameter, a researcher should not need to wait hours or minutes, but should receive instantaneous feedback -- which will measurably enhance their own cognitive performance~\citep{lu2022current}. We believe that flexible, interactive simulation with researcher-authored graphical user interface is not just a ``nice to have'', but a prerequisite for advanced robotics research.

	\section{Background} \label{background}

\lettrine[lines=2]{I}{n} this section we provide a general background on Optimal Control and Trajectory Optimisation, then focus on Predictive Control and derivative-free optimisation.

\paragraph{Optimal control.}
Optimal Control means choosing actions in order to minimise future costs (equivalently, to maximise future returns). A dynamical system with \emph{state} $\state \in \mathbf{R}^n$, which takes a user-chosen \emph{control} (or \emph{action}) $\control \in \mathbf{R}^m$, evolves according to the discrete-time\footnote{The continuous-time formulation is generally equivalent, we choose discrete time for notation simplicity.} dynamics:
\begin{equation}
    \nextstate = f(\state, \control),
    \label{dynamics}
\end{equation}
returning a new state, $\nextstate \in \mathbf{R}^n$. 
The behaviour of the system is encoded via the running cost: 
\begin{equation}
\cost(\state, \control),
\end{equation}
a function of state and control, where explicit time-dependence can be realised by folding time into the state. Future costs (a.k.a \emph{cost-to-go} or \emph{value}), can be defined in several ways. The summation can continue to infinity, leading to the \emph{average-cost} or \emph{discounted-cost} formulations, favoured in temporal-difference learning, which we discuss in Section \ref{discussion}. Here we focus on the \emph{finite-horizon} formulation, whereby the optimisation objective $\objective$ is given by:
\begin{equation}
    \objective(\state_{0:T}, \control_{0:T}) = \sum_{t=0}^{T} \cost(\state_t, \control_t), \label{objective}
\end{equation}
where subscripts indicated discrete-time indices.

\paragraph{Trajectory optimisation.}
Solving the finite-horizon optimal control problem (\ref{dynamics}, \ref{objective}), i.e., optimising a fixed-length trajectory, is commonly known as \emph{planning} or \emph{trajectory optimisation}. These algorithms \citep{von1992direct, betts1998survey} have a rich history reaching back to the Apollo Program \citep{manned1971apollo,smith1967trajectory}. An important distinction can be made between two classes of algorithms:
\begin{itemize}
\item \emph{Direct} or \emph{simultaneous} methods have both states and controls as decision variables and enforce the dynamics~\eqref{dynamics} as constraints. These methods (e.g., \cite{stryk1993numerical}) specify a large, sparse optimisation problem, which is usually solved with general-purpose software \citep{wachter2006implementation, gill2005snopt}. They have the important benefit that non-physically-realisable trajectories can be represented, for example in order to clamp a final state, without initially knowing how to get to it.
\item \emph{Shooting} methods like Differential Dynamic Programming \citep{jacobson1970differential} use only the controls $\control_{0:T}$ as decision variables and enforce the dynamics via forward simulation. In the shooting approach only physically-realisable trajectories can be considered, but they benefit from the reduced search space and from the optimiser not having to enforce the dynamics. The latter benefit is especially important for stiff systems like those with contact, where the difference between a physical and non-physical trajectory can be very small\footnote{For example, consider the physical scenario of a free rigid box lying flat on a plane under gravity, and then consider the non-physical scenario of the same box hovering above the plane or penetrating it by a few microns.}. Unlike \emph{direct} methods which require dynamics derivatives, shooting methods can employ derivative-free optimisation, as discussed below.
\end{itemize}
\paragraph{Predictive control.}

\begin{algorithm}[t]
	\caption{Predictive Control (asynchronous)}\label{predictive_control}
	\textbf{Agent}: (repeat)
	\begin{algorithmic}[1]
		\State Read the current action $\control$ from the \emph{nominal} plan $\plan$, apply it to the controlled system.
	\end{algorithmic}
	\textbf{Planner}: (repeat)
	\begin{algorithmic}[1]
		\State Measure the current state $\state$.
		\State Using the nominal $\plan$ to warm-start, optimise the finite-horizon objective $\objective$.
		\State Update $\plan$.
	\end{algorithmic}	
\end{algorithm}

The key idea of Predictive Control, invented in the late 60s and first published in \citep{richalet1978model}, is to use trajectory optimisation in \emph{real-time} as the system dynamics are evolving. This class of algorithm has been successfully deployed in numerous real-world settings including: chemical and nuclear process control \citep{na2003model, lopez2013fast}, navigation for autonomous vehicles \citep{falcone2007predictive}, and whole-body control of humanoid robots \citep{kuindersma2016optimization}.

In the real-time setting, the current state $\state$ needs to be estimated or measured, and the trajectory optimiser is required to return a set of optimal or near-optimal controls for the finite-horizon (here often called the \emph{receding horizon}) problem, starting at $\state$. We use $\plan$ to denote the \emph{plan}, the finite-horizon policy. In the context of shooting methods $\plan = \control_{0:T}$, though as we discuss later, in some cases it can be re-parameterised, rather than using the discrete-time control sequence directly. Predictive control is best thought of in terms of two asynchronous processes, the \emph{agent} and the \emph{planner}, see Algorithm \ref{predictive_control}.

Predictive Control has the following notable properties:
\begin{itemize}
\item Faster computation improves performance. The reason for this is clear, the more optimisation steps the planner can take in one unit of time, the better the optimised control sequences will be.
\item Warmstarting has a large beneficial effect. By reusing the plan from the previous planning step, the optimiser only needs to make small modifications in order to correct for the changes implied by the new state. Warm-starting also leads to an amortisation of the optimisation process across multiple planning steps.
\item The optimisation is not required to converge, only to improve, see Section \ref{ps_discussion}.
\item Tasks with a behaviour timescale much longer than the planning horizon $T$ are often still solvable, though this property is task dependent.
\item Predictive controllers can easily get stuck in local minima, especially those using derivatives. This is due to the myopic nature of the optimisation, and can be addressed by terminating the rollout with a value function approximation, see Section \ref{future}.
\end{itemize}

\paragraph{Derivative-free optimisation.}
One of the main reasons for our initial focus on shooting methods is their ability to make use of derivative-free optimisation, also known as \emph{sampling-based} optimisation~\citep{audet2017derivative}. Despite not leveraging problem structure or gradient information, this class of algorithms can discover complex behaviours \citep{salimans2017evolution, mania2018simple}.

Sampling-based methods maintain a search distribution over policy parameters and evaluate the objective at sampled points in order to find an improved solution. Popular algorithms include: random search \citep{matyas1965random}, genetic algorithms \citep{holland1992genetic}, and evolutionary strategies \citep{rechenberg1973evolutionsstrategie}, including CMA-ES \citep{hansen2001completely}. 

This class of algorithms has a number of desirable properties. First, because derivative information is not required, they are well-suited for tasks with non-smooth and discontinuous dynamics. Second, these methods are trivially parallelisable.

Sampling-based methods have been used in the predictive control context, but not very widely. Notable exceptions are H{\"a}m{\"a}l{\"a}inen's work \citep{hmlinen2014online, hmlinen2015online}, which is an early precursor to \app{}, and the oeuvre of Theodorou including~\citep{williams2017model} and related papers. While these methods are usually considered to be sample inefficient, we explain below why, specifically in the predictive control context, they can be surprisingly competitive.

	\section{MuJoCo MPC (\app{})} \label{trajectory_explorer}

\lettrine[lines=2]{W}{e} introduce \app{}, an open-source interactive application and software framework for predictive control, that lets the user easily synthesise behaviours for complex systems using predictive control algorithms in real time. Behaviours are specified by simple, composable objectives that are risk-aware. The planners, including: Gradient Descent, Iterative Linear Quadratic Gaussian (iLQG), and Predictive Sampling are implemented in C++ and extensively utilise multi-threading for parallel rollouts. The framework is asynchronous, enabling simulation slow-down and emulation of a faster controller, allowing this tool to run on slow machines. An intuitive graphical user-interface enables real-time interactions with the environment and the ability to modify task parameters, planner and model settings, and to instantly see the effects of the modifications. The tool is available at: \colorhref{https://github.com/deepmind/mujoco_mpc}{https://github.com/deepmind/mujoco\_mpc}.

\subsection{Physics Simulation}
We build \app{} using the API of the open-source physics engine MuJoCo \citep{todorov2012mujoco}.  MuJoCo is a good infrastructure for an interactive framework for robotics algorithms for two main reasons: first, MuJoCo supports simulating multiple candidate future trajectories in parallel by offering a thread-safe API, which maximises the utilisation of modern multi-core CPU architectures; second, MuJoCo affords faster-than-realtime simulation of high-dimensional systems with many contacts --- for example, the humanoid (a 27-DoF system) can be simulated 4000 times faster than realtime on a single CPU thread.

\subsection{Objective}
\app{} provides convenient utilities to easily design and compose costs in order to specify an objective \eqref{objective}; as well as automatically and efficiently compute derivatives.

\paragraph{Costs.}

We use a ``base cost'' of the form:
\begin{equation}
\label{eq:riskfreecost}
\precost(\state, \control) = \sum_{i = 0}^M w_i \cdot \text{n}_i\big(\residual_i(\state, \control)\big).
\end{equation}
This cost is a sum of $M$ terms, each comprising:
\begin{itemize}
\item A nonnegative weight $w \in \mathbf{R}_{+}$ determining the relative importance of this term.
\item A twice-differentiable norm $\text{n}(\cdot) : \mathbf{R}^p \rightarrow \mathbf{R}_+$, which takes its minimum at $0^p$.
\item The residual $\residual \in \mathbf{R}^p$ is a vector of elements that are ``small when the task is solved''.
\end{itemize}

\paragraph{Risk sensitivity.}

\begin{figure}[t]
	\centering
	\includegraphics[height=7.0cm]{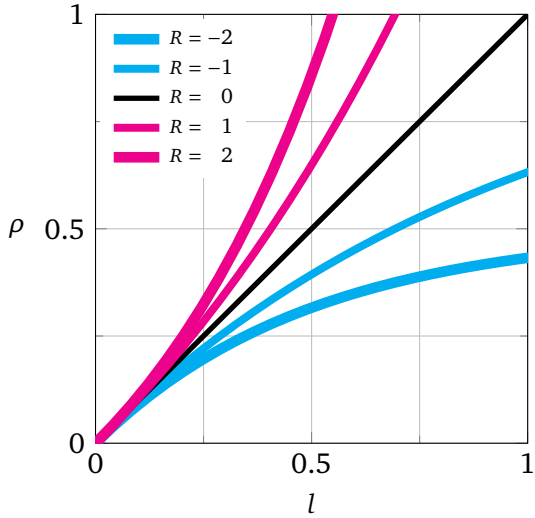}
	\caption{Risk transformation $\rho(l ; R)$. The function is evaluated between $0$ and $1$ for different values of the risk parameter $R$.}
	\label{fig:risk}
\end{figure}

We augment the base cost \eqref{eq:riskfreecost} with a risk-aware exponential scalar transformation, $\rho : \mathbf{R}_+ \times \mathbf{R} \rightarrow \mathbf{R}$, corresponding to the classical risk-sensitive control framework \citep{jacobson1973optimal, whittle1981risk}. The final running cost $\cost$ is given by:
\begin{equation}
\cost(\state, \control) = \rho(\precost(\state, \control) ; R) = \frac{e^{R\cdot \precost(\state, \control)} - 1}{R}
\label{eq:risk}
\end{equation}
The scalar parameter $R \in \mathbf{R}$ denotes risk-sensitivity. $R=0$ (the default) is interpreted as risk-neutral, $R>0$ as risk-averse, and $R<0$ as risk-seeking.
The mapping $\rho$ (see Figure \ref{fig:risk}) has the following properties:

\begin{itemize}
\item Defined and smooth for any $R$.
\item If $R=0$, is the identity $\rho(\precost; 0) = \precost$ (in the limit).
\item Zero is a fixed point: $\rho(0; R) = 0$.
\item The derivative at 0 is 1: $\partial \rho(0; R) / \partial \precost = 1$.
\item Non-negative: If $\precost \geq 0$ then $\rho(\precost; R) \geq 0$.
\item Monotonic: $\rho(\precost; R) > \rho(z; R)$ if $\precost > z$.
\item If $R<0$ then $\rho$ is bounded: $\rho(\precost; R) < - \frac{1}{R}$.
\item $\rho(\precost;R)$ carries the same units as $\precost$.
\end{itemize}

Note that for negative $R$, the transformation $\rho$ creates costs that are similar to the bounded rewards commonly used in reinforcement learning. For example, when using the quadratic norm $\text{n}(r) = r^T W r$ for some SPD matrix $W=\Sigma^{-1}$ and a risk parameter $R=-1$, we get an inverted-Gaussian cost $\cost=1 - e^{-r^T \Sigma^{-1} r}$, whose minimisation is equivalent to maximum-likelihood maximisation of the Gaussian. This leads to the interesting interpretation of the bounded rewards commonly used in RL as \emph{risk-seeking}. We do not investigate this relationship further in this paper.

\paragraph{Derivatives.}
MuJoCo provides a utility for computing finite-difference (FD) Jacobians of the dynamics, which is efficient in two ways. First, by avoiding re-computation where possible, for example when differencing w.r.t. controls, quantities that depend only on positions and velocities are not recomputed. Second, because FD computational costs scale with the dimension of the \emph{input}, outputs can be added cheaply. MuJoCo's step function $\nextstate, \residual = f(\state, \control)$, computes both the next state $\nextstate$ and sensor values $r$, defined in the model. Because the FD approximation of the Jacobians,
\begin{equation}
    \frac{\partial \nextstate}{\partial \state}, \frac{\partial \nextstate}{\partial \control}, \frac{\partial \residual}{\partial \state}, \frac{\partial \residual}{\partial \control},
\end{equation}
scales like the combined dimension of $\state$ and $\control$, adding more sensors $\residual$ is effectively ``free''. \app{} automatically and efficiently computes cost derivatives as follows.
\paragraph{Gradients.} Cost gradients are computed with:
\begin{subequations}
\begin{align}
    \frac{\partial \cost}{\partial \state} &= e^{R \precost} \frac{\partial \precost}{\partial \state}=e^{R \precost}\sum \limits_{i = 0}^M w_i \frac{\partial \text{n}_i}{\partial r} \frac{\partial r_i}{\partial \state}, \\
    \frac{\partial \cost}{\partial \control} &= e^{R \precost} \frac{\partial \precost}{\partial \control} = e^{R \precost}\sum \limits_{i = 0}^M w_i \frac{\partial \text{n}_i}{\partial r} \frac{\partial r_i}{\partial \control}.
\end{align}
\end{subequations}
The norm gradients, $\partial \text{n} / \partial r$, are computed analytically.

\paragraph{Hessians.}

Second-order derivatives use the Gauss-Newton approximation, ignoring second derivatives of $\residual$:
\begin{subequations}
\begin{align}
    \frac{\partial^2 \cost}{\partial \state^2} \approx&\,  e^{R \precost} \left[ \sum \limits_{i = 0}^M w_i {\frac{\partial r_i}{\partial \state}}^T\frac{\partial^2 \text{n}_i}{\partial r^2} \frac{\partial r_i}{\partial \state} + R {\frac{\partial \precost}{\partial \state}}^T \frac{\partial \precost}{\partial \state} \right], \\
    \frac{\partial^2 \cost}{\partial \control^2} \approx& \, e^{R \precost} \left[ \sum \limits_{i = 0}^M w_i {\frac{\partial r_i}{\partial \control}}^T\frac{\partial^2 \text{n}_i}{\partial r^2} \frac{\partial r_i}{\partial \control}  + R {\frac{\partial \precost}{\partial \control}}^T \frac{\partial \precost}{\partial \control} \right],\\
    \frac{\partial^2 \cost}{\partial \state \partial \control} \approx& \, e^{R \precost} \left[ \sum \limits_{i = 0}^M w_i {\frac{\partial r_i}{\partial \state}}^T \frac{\partial^2 \text{n}_i}{\partial r^2} \frac{\partial r_i}{\partial \control} + R {\frac{\partial \precost}{\partial \state}}^T \frac{\partial \precost}{\partial \control} \right].
\end{align}
\end{subequations}
The norm Hessians, $\partial^2 \text{n}/\partial r^2$, are computed analytically.


\subsection{Splines}

\begin{figure}[t]
	\centering
	\includegraphics[width=0.95\columnwidth,height=6.5cm]{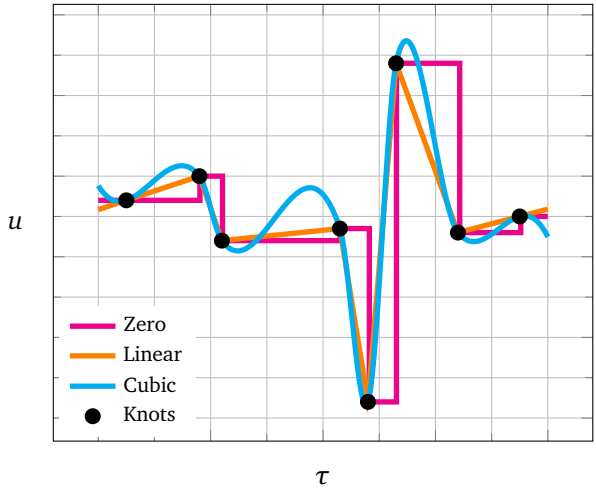}
	\caption{Time-indexed spline representation of the controls. Parameter points (black) utilised to construct: zero (magenta), linear (orange), and cubic (blue) interpolants.}
	\label{spline}
\end{figure}

As we explain below, planners like iLQG require the \emph{direct} control-sequence representation $\control_{0:T}$ due to the requirements of the Bellman Principle. Without this constraint, controls can be ``compressed'' into a lower-dimensional object. There are many ways to do this, we picked the simplest: splines. Action trajectories are represented as a time-indexed set of knots, or control-points, parameterised by a sequence of monotonic time points $\tau_{0:P}$ and parameter values $\theta_{0:P}$, where we use the shorthand $\theta = \theta_{0:P}$. Given a query point $\tau$, the evaluation of the spline is given by:
\begin{equation}
\control = s\big(\tau; (\tau_{0:P}, \theta) \big).
\end{equation}
We provide three spline implementations: traditional cubic Hermite splines, piecewise-linear interpolation, and zero-order hold. See (Fig. \ref{spline}) for an illustration.

The main benefit of compressed representations like splines is that they reduce the search space. They also smooth the control trajectory, which is often desirable. Spline functions belong to the class of linear bases, which includes the Fourier basis and orthogonal polynomials. These are useful because they allow easy propagation of gradients from the direct representation $\partial \plan$  back to the parameter values $\partial \theta$. In our case this amounts to computing:
\begin{equation}\label{eq:splinederiv}
\frac{\partial s}{\partial \theta},
\end{equation}
which has a simple analytic formula (see code for details). Unlike other linear bases, splines have the convenient property that bounding the values $\theta$ also bounds the spline trajectory. This is exactly true for the zero and linear interpolations, and mostly-true for cubic splines. Bounding is important as most physical systems clamp controls to bounds, and there is no point searching outside of them. Expressions for cubic, linear and zero interpolations are provided in Appendix A.

\subsection{Planners}
\app{} includes two derivative-based planners. 

\paragraph{iLQG.}

\begin{algorithm}[t]
	\caption{iLQG}\label{ilqg_algorithm}
	\begin{algorithmic}[1]
	    \Require initial state $\state_0$, nominal plan $\plan=\control_{0:T}$
	    \State Roll out nominal trajectory from $\state_0$ using $\plan$
		\State Compute action improvements and feedback policy using Dynamic Programming.
		\State Roll out parallel line search with feedback policy \eqref{ilqg_feedback_policy}.
		\State Best actions are new nominal actions.
	\end{algorithmic}
\end{algorithm}

The iLQG\footnote{Equivalently, ``iLQR'', since we don't make use of the noise-sensitive term for which iLQG was originally developed~\citep{li2004iterative}. We keep the name ``iLQG'' due to its provenance.} planner~\citep{tassa2012synthesis}, a Gauss-Newton approximation of the DDP algorithm~\citep{jacobson1970differential}, utilises first- and second-order derivative information to take an approximate Newton step over the open-loop control sequence $\control_{0:T}$ via dynamic programming \citep{kalman1964lqr}, producing a time-varying linear feedback policy:
\begin{equation}
    \control_t = \bar{\control}_t + K_t (\state_t - \bar{\state}_t) + \alpha k_t \label{ilqg_feedback_policy}.
\end{equation}
The \emph{nominal}, or current best trajectory, is denoted with overbars ($\bar{\phantom{x}}$), $K$ is a feedback gain matrix, and $k$ is an improvement to the current action trajectory. A parallel line search over the step size $\alpha \in [\alpha_{\textrm{min}}, 1]$ is performed to find the best improvement. Additional enhancements include a constrained backward pass \citep{tassa2014control} that enforces action limits and adaptive regularisation. The details of iLQG are too involved to restate here, we refer the reader to the references above for details.

\paragraph{Gradient descent.}

\begin{algorithm}[t]
	\caption{Gradient Descent}\label{gradient_descent_algorithm}
	\begin{algorithmic}[1]
	    \Require initial state $\state_0$, nominal plan $\plan(\theta)$
		\State Roll out nominal from $\state_0$ using $\plan(\theta)$
		\State Compute $\partial \objective/\partial \plan$ with \eqref{eq:pontryagin}
		\State Compute $\partial \objective/\partial \theta$ with \eqref{eq:paramderivs}
		\State Roll out parallel line-search with \eqref{eq:linesearch}
		\State Pick the best one: $\theta \leftarrow \mbox{argmin} \bigl(\objective(\theta^{(i)})\bigr)$
	\end{algorithmic}
\end{algorithm}

This first-order planner, known as Pontryagin's Maximum Principle \citep{mangasarian1966sufficient}, utilises gradient information to improve action sequences, here represented as splines. The gradient of the total return is used to update the spline parameters, using a parallel line search over the step size $\alpha \in [\alpha_{\textrm{min}}, \alpha_{\textrm{max}}]$:
\begin{equation} \label{eq:linesearch}
    \theta \leftarrow \theta - \alpha \frac{\partial J}{\partial \theta}.
\end{equation}
The total gradient is given by:
\begin{equation}\label{eq:paramderivs}
    \frac{\partial J}{\partial \theta} = \frac{\partial J}{\partial \plan} \frac{\partial \plan}{\partial \theta},
\end{equation}
where the spline gradient $\partial \plan/\partial \theta$ is given by \eqref{eq:splinederiv}, while $\partial J/\partial \plan$ is computed with the Maximum Principle. Letting $\lambda$ denote the \emph{co-state}, the gradients with respect to $\control$ are given by:
\begin{subequations}\label{eq:pontryagin}
\begin{align}
    \lambda_{j} &= \frac{\partial c}{\partial \state_j} + \Bigl(\frac{\partial f}{\partial \state_j}\Bigr)^T\lambda_{j+1}, \\
    \frac{\partial \objective}{\partial \control_j} &= \frac{\partial c}{\partial \control_j} + \Bigl(\frac{\partial f}{\partial \control_j}\Bigr)^T\lambda_{j+1}.
\end{align}
\end{subequations}
The primary advantage of this first-order method compared to a computationally more expensive method like iLQG is that optimisation is performed over the smaller space of spline parameters, instead of the entire (non-parametric) sequences of actions.

\paragraph{Predictive Sampling.}

\begin{algorithm}[t]
	\caption{Predictive Sampling}\label{predictive_sampling_algorithm}
	\textbf{Parameters:} $N$ rollouts, noise scale $\sigma$
	\begin{algorithmic}[1]
	    \Require initial state $\state_0$, nominal plan $\plan(\theta)$
	    \State Get $N\!-\!1$ samples $\theta_i \sim \mathcal{N}(\theta, \sigma^2)$
		\State Including $\theta$, roll out all $N$ samples from $\state_0$
        \State Pick the best one: $\theta \leftarrow \mbox{argmin} \bigl(\objective(\theta^{(i)})\bigr)$
	\end{algorithmic}
\end{algorithm}

This is a trivial, zero-order, sampling-based Predictive Control method that works well and is easy to understand. Designed as an elementary baseline, this algorithm turned out to be surprisingly competitive with the more elaborate derivative-based algorithms.

\paragraph{Algorithm.} A nominal sequence of actions, represented with spline parameters, is iteratively improved using random search \citep{matyas1965random}. At each iteration, $N$ candidate splines are evaluated: the nominal itself and $N-1$ noisy samples from a Gaussian with the nominal as its mean and fixed standard deviation $\sigma$. After sampling, the actions are clamped to the control limits by clamping the spline parameters $\theta$. Each candidate's total return is evaluated and the nominal is updated with the best candidate. See Algorithm \ref{predictive_sampling_algorithm} and pseudocode in Appendix C. Predictive Sampling is not innovative or performant, but is presented as a simple baseline, see Discussion below.


	\section{Results} \label{results}

\lettrine[lines=2]{W}{e} provide a short textual description of our graphical user interface (GUI) for three example tasks. They are best understood by viewing the associated video at \colorhref{https://dpmd.ai/mjpc}{dpmd.ai/mjpc} or better yet, by downloading the software and interacting with it. 

\subsection{Graphical User Interface}

\begin{figure*}[t]
	\centering
	\includegraphics[width=.95\textwidth]{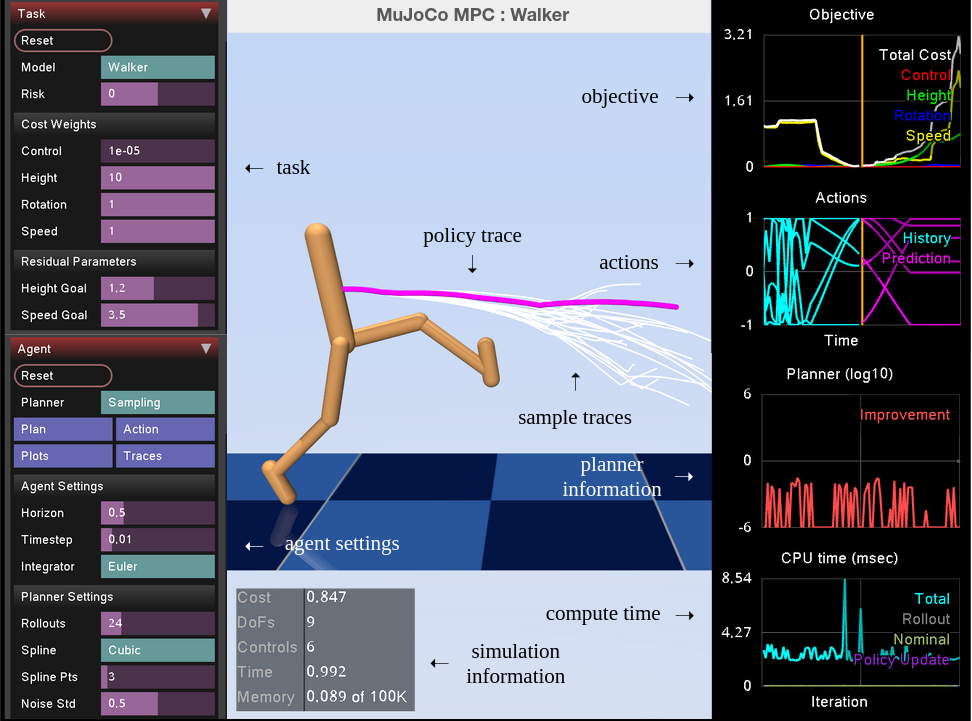}%
	\caption{Graphical User Interface. The left tab includes modules for Tasks and the Agent. In the Task module, models are selected from a drop-down menu and the risk value is set; cost weights and residual parameters are specified with interactive sliders. The Agent module provides access to a drop-down menu for planners, settings, and switches that toggle the planner and controller. The right tab includes simulation readings and predictions for the cost, including individual terms, and actions. Live, planner-specific information and various compute time results are shown below. Traces of the current policy and sampled trajectories are visualised and the user can interactively pause or slow down the simulation and apply external forces and moments to the system.}%
	\label{fig:gui}
\end{figure*}

The MJPC GUI, shown and described in Fig. \ref{fig:gui}, provides an interactive simulation environment, as well as modules containing live plots and parameters that can be set by the researcher. The intuitive interface makes policy design easy by enabling the researcher to interactively change cost parameters or planner settings and immediately see the results in both the simulation environment and live plots, allowing fast debugging and an enhanced understanding of the factors that influence behaviour.

\subsection{Examples}

\begin{figure*}[h] 
	\captionsetup[subfigure]{justification=centering}
	\begin{subfigure}{0.99\textwidth}
		\centering
		\includegraphics[width=.175\textwidth]{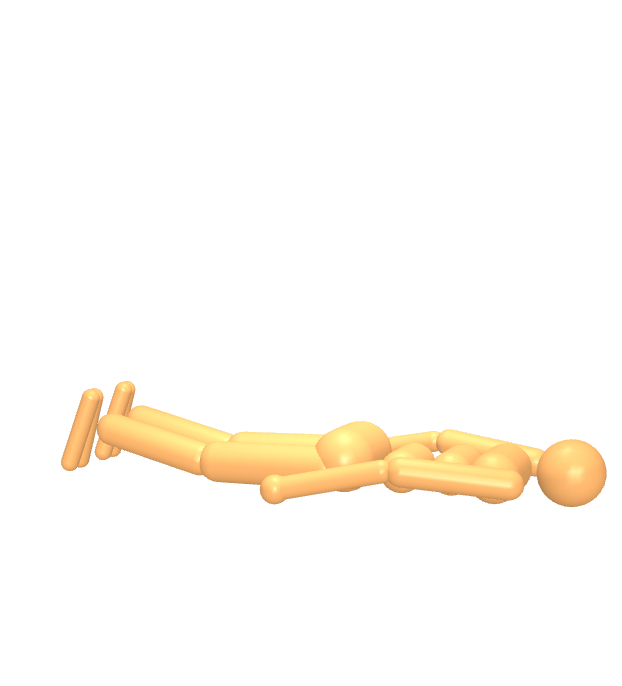}%
		\includegraphics[width=.175\textwidth]{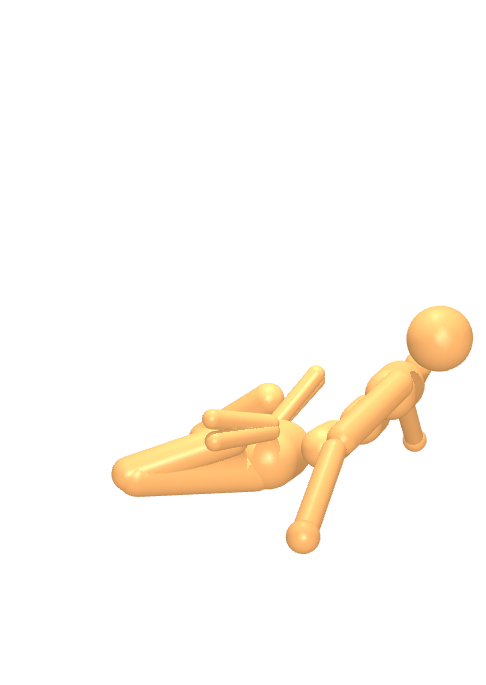}%
		\includegraphics[width=.175\textwidth]{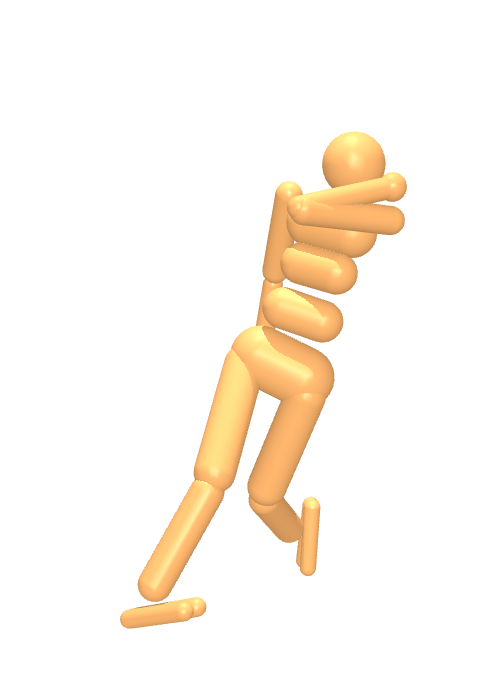}%
		\includegraphics[width=.175\textwidth]{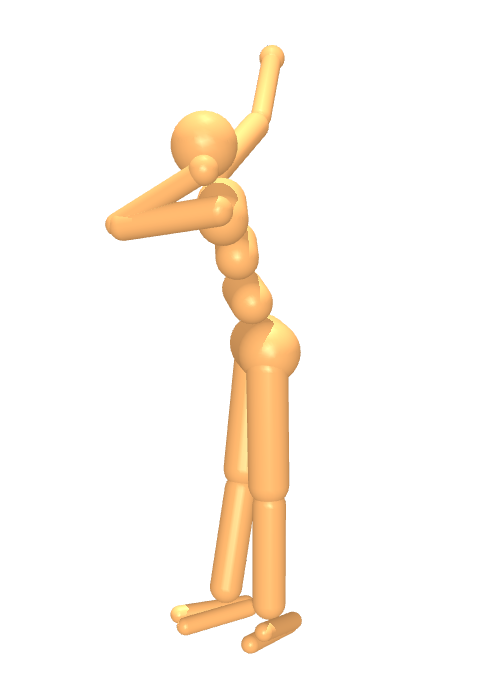}%
		\includegraphics[width=.175\textwidth]{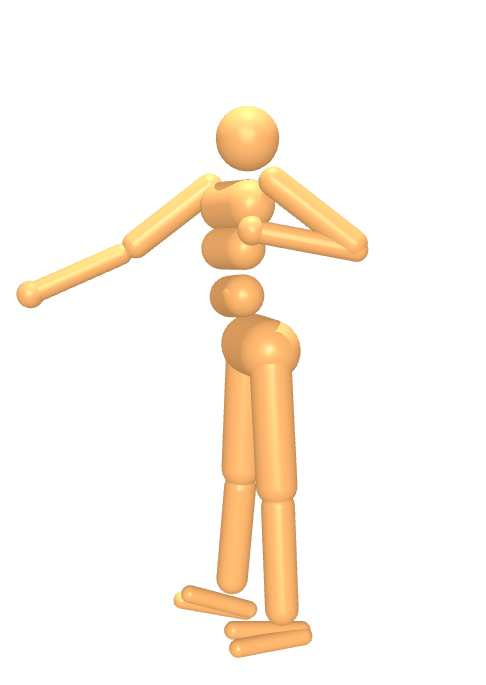}%
		\caption{Humanoid standing up off the floor.}
		\label{fig:humanoid_standing}
		\vspace*{7.5mm} 
	\end{subfigure}
	\begin{subfigure}{0.99\textwidth}
		\centering
		\includegraphics[width=.175\textwidth]{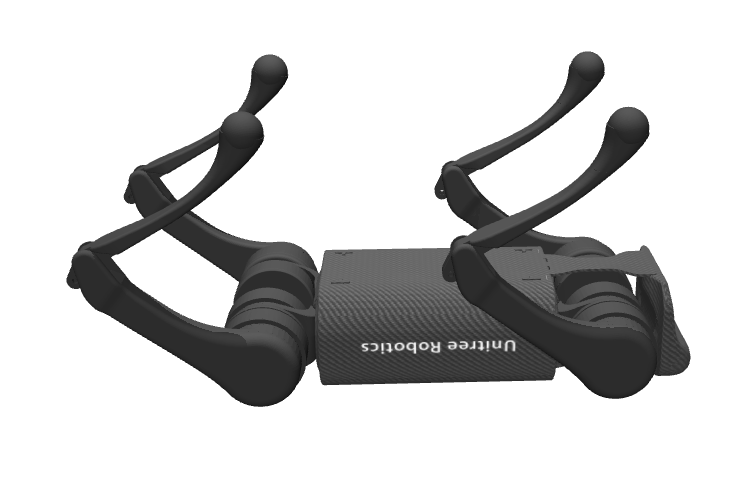}%
		\includegraphics[width=.175\textwidth]{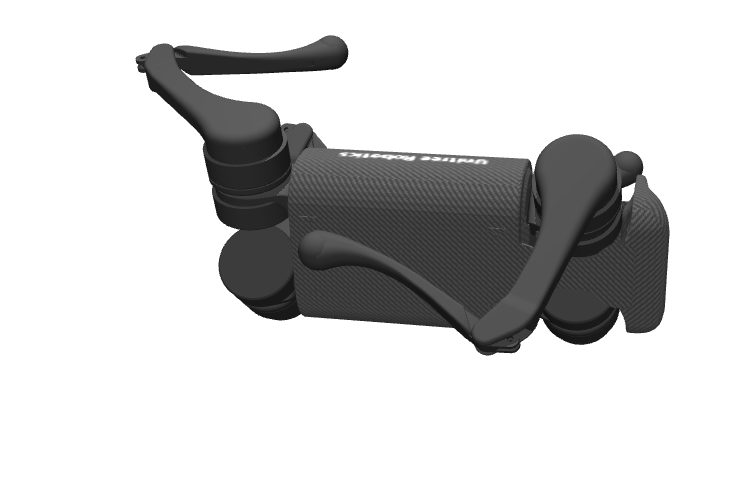}%
		\includegraphics[width=.175\textwidth]{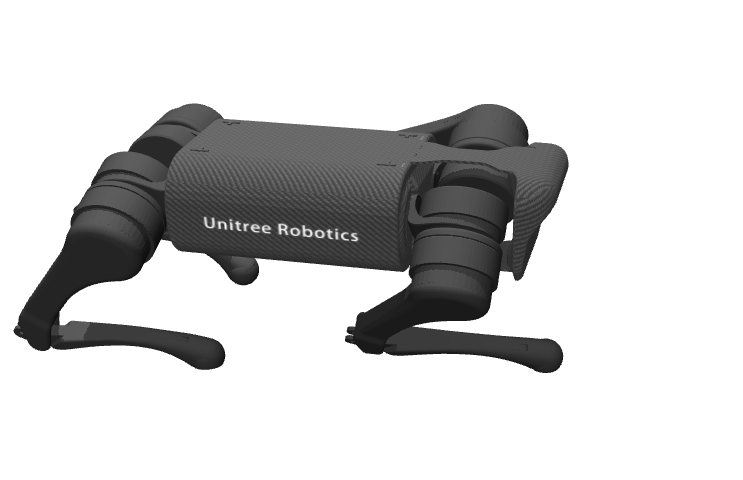}%
		\includegraphics[width=.175\textwidth]{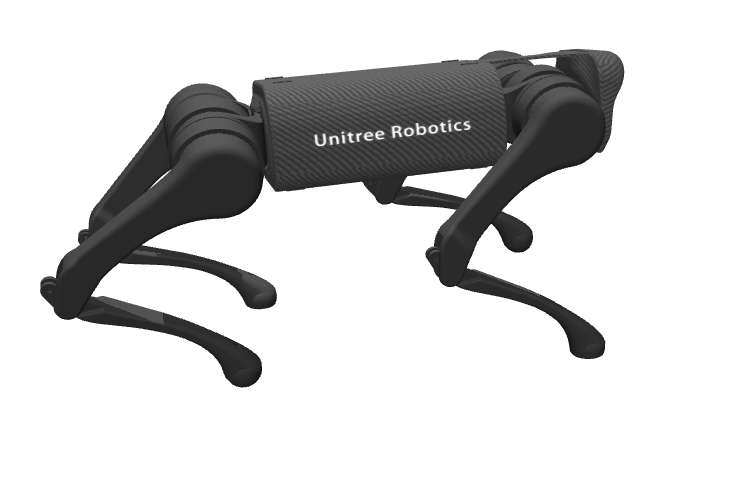}%
		\includegraphics[width=.175\textwidth]{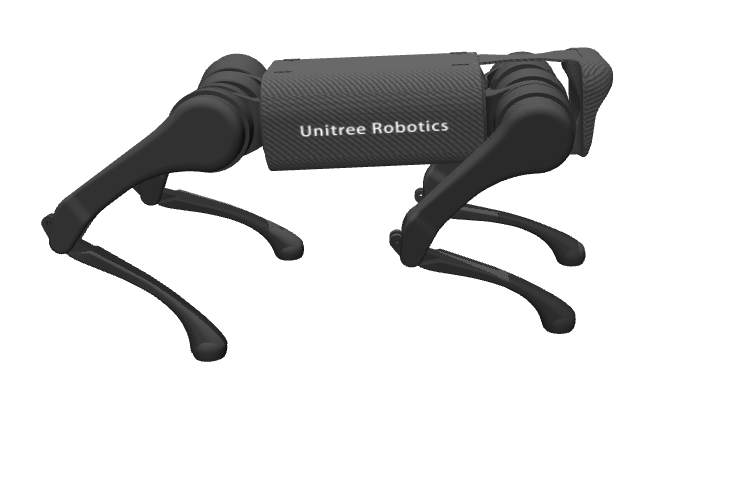}%
		\caption{Quadruped rolling off its back to stand up.}
		\label{fig:quadruped}
		\vspace*{7.5mm} 
	\end{subfigure}
	\begin{subfigure}{0.98\textwidth}
		\centering
		\includegraphics[width=.225\textwidth]{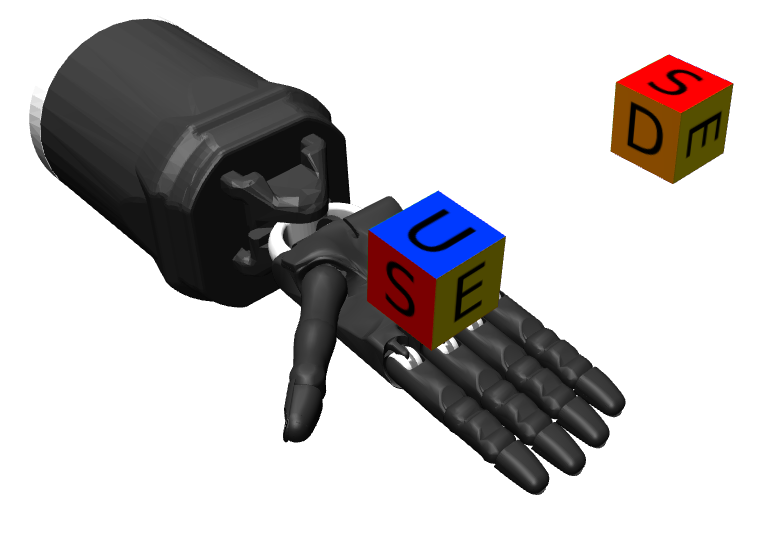}%
		\includegraphics[width=.225\textwidth]{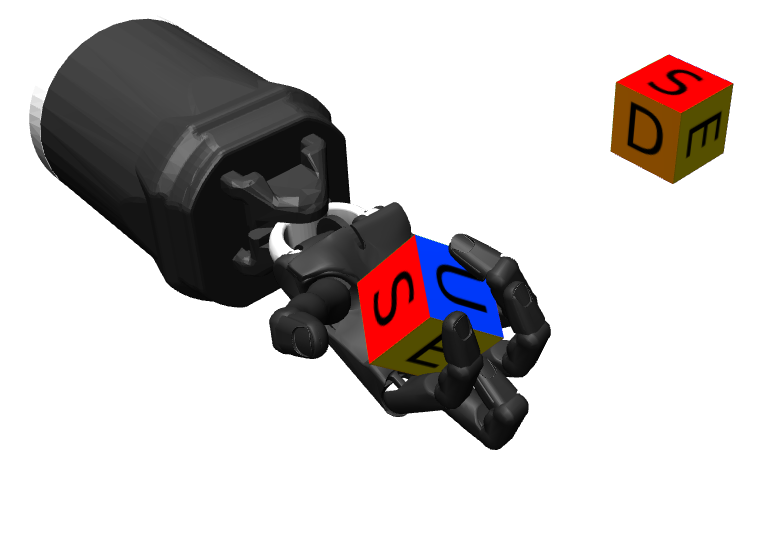}%
		\includegraphics[width=.225\textwidth]{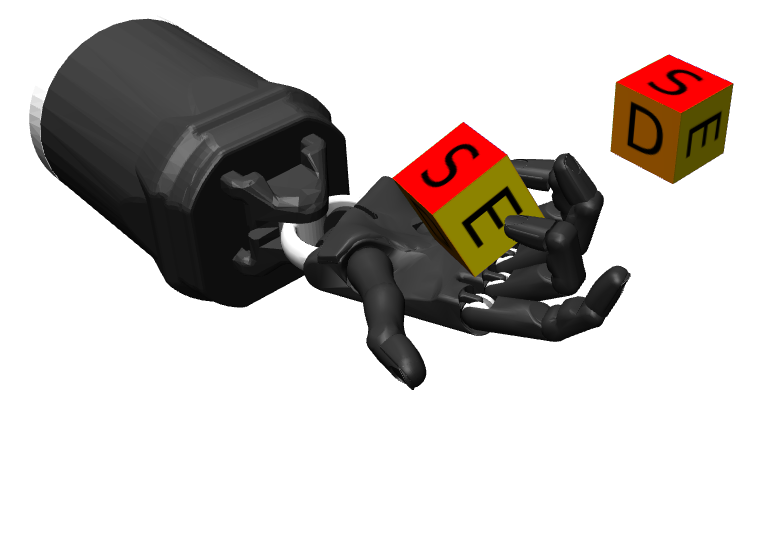}%
		\includegraphics[width=.225\textwidth]{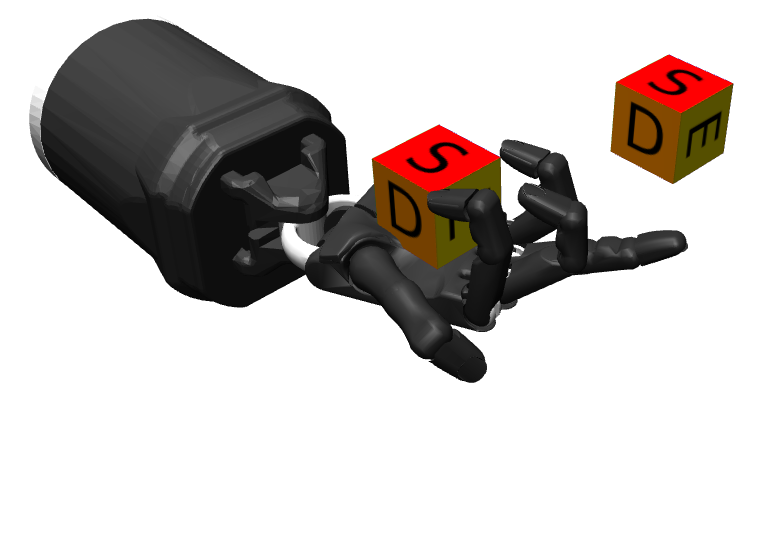}%
		\caption{Hand manipulating a cube to a goal orientation.}
		\label{fig:hand}
	\end{subfigure}
	\caption{Behaviours generated with MuJoCo MPC. Time progresses left to right.}
	\label{fig:examples}
\end{figure*}

In the following examples, we demonstrate the ability to synthesise complex locomotion and manipulation behaviours for a variety of high-dimensional systems in simulation on a single CPU. Further, we demonstrate that the behaviours are robust to disturbances and mismatch between the simulation and planning model, and can adapt extremely quickly in new scenarios. For all of the examples, the total planning time for a single update is between 1 and 20 milliseconds. We highlight three examples below and provide additional examples with the software. Experimental details for objectives and planner settings are found in the Appendix B.

\paragraph{Humanoid.}
This 27-DOF human-like system, from DeepMind Control Suite \citep{tunyasuvunakool2020dm_control}, has 21 actions and is tasked with standing. The system can be initialised on the floor and quickly stands in a manner that is robust to large disturbances. If a sufficiently large disturbance knocks the humanoid onto the floor, the system will stand back up (Fig. \ref{fig:humanoid_standing}).

\paragraph{Quadruped.}
A Unitree A1 quadruped \citep{unitree2022a1}, from MuJoCo Menagerie \citep{menagerie2022github}, exhibits agile behaviour to traverse uneven terrain which includes walking over a steep slope. On slower machines, the quadruped often struggles to ascend. In this scenario, the simulation slow down can be effectively utilised to provide the planner with addition simulation time to plan a successful climb. The system is also capable of rolling off its back and standing up (Fig. \ref{fig:quadruped}). In order to perform long-horizon tasks like continuously navigating the terrain, a series of target poses are set. Once a goal is reached, an automatic transition occurs and the next target is set. 

\paragraph{Hand.}
A Shadow Hand \citep{tuffield2003shadow}, also from MuJoCo Menagerie, performs in-hand manipulation of a cube to a desired orientation (Fig. \ref{fig:hand}), where this goal can be set by the researcher in real-time by interactively setting the target orientation. In-hand reorientation---a high-DoF system with complex contact dynamics---is considered difficult to solve~\citep{chen2022system} and to the best of our knowledge has not previously been solved from scratch, in real time.

	\section{Discussion} \label{discussion}
\lettrine[lines=2]{T}{he} thrust of this paper is to make predictive control accessible via customisable, interactive, open-source tooling. We believe that responsive, GUI-based tools are a prerequisite for accelerated robotics research, and that due to their importance, these tools should be modifiable and the inner workings transparent to the researcher. We hope that our \app{} project will be embraced by the community, and look forward to improving and extending it together.
 
\subsection{Predictive Sampling} \label{ps_discussion}
The effectiveness of this simple method suggests that fast, approximate optimisation can be competitive with more sophisticated methods which return better solutions but at a lower rate. Does the higher planning rate completely explain this surprising effectiveness? We believe there is another, subtler reason. Predictive Control is not well-described by the tenets of traditional optimisation. For example, it usually makes no sense to take more than one step of optimisation. Once a single iteration is complete, it is more important to measure a new value of the state and re-plan, than it is to continue to converge to the minimum of an already-outdated problem. The constant shifting of the optimisation landscape makes Predictive Control a \emph{qualitatively different problem}, more like surfing than mountain climbing. The goal is not to find the minimum, but to \emph{remain in the basin-of-attraction of the minimum}. This is a different, weaker criterion, at which simple algorithms fair better than when measured by the traditional yardstick of convergence.

To be clear, Predictive Sampling is not a novel algorithm; instead, it is a baseline. A corner-case of many existing methods, it can variously be described as ``MPPI with infinite temperature'',  ``CEM with a non-adaptive distribution'' or just ``trivial random search''. Better algorithms exist, but none are so easy to describe or implement. We are introducing Predictive Sampling not because it is good, but because it is \emph{not good}. It is the simplest possible sampling-based shooting method, and therefore establishes a \emph{lower bound} for performance baselines.

\subsection{Use cases} \label{current}

Before we discuss limitations and their possible resolutions, it is worth asking how can \app{} be used \emph{now}, as it is described above?
\begin{enumerate}
    \item \textbf{Task design}. \app{} makes it easy to add new tasks, expose task parameters to the GUI, and quickly generate the desired behaviour. The task can then be re-implemented in any other framework of choice. While we have not yet implemented time-dependent tasks, it is possible and easy; we expect \app{} to work especially well for motion-tracking tasks.
    \item \textbf{Data generation}. \app{} can be used to generate data for learning-based approaches, i.e., it can act like an ``expert policy''. In this context, it is often the case that the model and task from which the data is generated do not have to exactly match the one used by the learner, and the data can likely be useful for a wide range of setups.
    \item \textbf{Predictive Control research}. For researchers interested in Predictive Control itself, \app{} provides an ideal playground. \app{} can switch planners on-the-fly, and its asynchronous design affords a fair comparison by correctly accounting for and rewarding faster planners.
\end{enumerate}

\subsection{Limitation and Future Work} \label{future}

\paragraph{Can only control what MuJoCo can simulate.} This is a general limitation of Predictive Control and is in fact stronger since one can only control what can be simulated \emph{much faster than real-time}. For example, it is difficult to imagine any simulation of a very-high-DoF system, like fluid, cloth or soft bodies advancing so fast. One solution is improved simulation using a combination of traditional physics modeling and learning, e.g., \citep{ladicky2017physics}. Another possibility is to entirely learn the dynamics from observations. This approach, often termed Model Based Reinforcement Learning is showing great promise \citep{heess2015learning, nagabandi2020deep, wu2022daydreamer}. We would recommend that where possible, when attempting Predictive Control using learned dynamics models, a traditional simulator be employed as a fallback as in \citep{schrittwieser2020mastering}, to disambiguate the effects of modeling errors. 

\paragraph{Myopic.} The core limitation of Predictive Control is that it is \emph{myopic} and cannot see past the fixed horizon. This can be ameliorated in three conceptually straightforward ways:
\begin{enumerate}
    \item \textbf{Learned policies}. By adding a learned policy, information from past episodes can propagate to the present via policy generalisation~\citep{byravan2021evaluating}. This approach is attractive since it can only \emph{improve} performance: when rolling out samples, one also rolls out the proposal policy. If the rollout is better, it becomes the new nominal. A learned policy is also expected to lead to more stereotypical, periodic behaviours, which are important in locomotion. 
    \item \textbf{Value functions}. Terminating the rollout with a learned value function which estimates the remaining cost-to-go is the obvious way by which to increase the effective horizon. Combining learned policies and value functions with model-based search would amount to an ``AlphaGo for control''~\citep{silver2016mastering, springenberg2020local}.
    \item \textbf{High-level agent}. A predictive controller could be used as the low-level module in a hierarchical control setup. In this scenario, the actions of the high-level agent have the semantics of setting the cost function of the predictive controller. The predictive controller remains myopic while the high-level agent contains the long-horizon ``cognitive'' aspects of the task. A benefit of this scenario is that the high-level actions have a much lower frequency than required for low-level control (e.g., torques).
\end{enumerate}
\paragraph{Hardware.} \app{} is aimed at robotics research, which raises the question, can it be used to control hardware?
\begin{enumerate}
    \item \textbf{Transfer learning}. As mentioned in \ref{current}, using \app{} to generate data which can then be transferred to a real robot is already possible.
    \item \textbf{Estimation}. The most obvious yet difficult route to controlling hardware is to follow in the footsteps of classic control and couple \app{} to an estimator providing real-time state estimates. In the rare cases where estimation is easy, for example with fixed-base manipulators and static objects, controlling a robot directly with \app{} would be a straightforward exercise. The difficult and interesting case involves free-moving bodies and contacts, as in locomotion and manipulation. For certain specific cases, like locomotion on flat, uniform terrain, reasonable estimates should not be difficult to obtain. For the general case, we believe that contact-aware estimation is possible following the approach of \citep{lowrey2014physically}, but that remains to be seen. Relatedly, we believe that high-quality estimators also require the same kind of interactive, GUI-driven interface used by \app{}.
\end{enumerate}

	\bibliography{references}

	\begin{appendices}
		\section{Interpolation}
This section provides analytical expressions for computing interpolations. The indices $j$ and $j+1$ correspond to the indices of the domain variables which contain the query point $\tau$. These values are efficiently found using binary search in $\mathbf{O}\left(\text{log}(n)\right)$ where $n$ is the dimension of the domain variable set.

\paragraph{Zero.}
The zero-order interpolation simply returns the parameter values at the lower bound index:
\begin{equation}
    \theta_j \leftarrow s.
\end{equation}

\paragraph{Linear.}
The linear interpolation returns: 
\begin{equation}
    (1 - q) \cdot \theta_j + q \cdot \theta_{j+1} \leftarrow s,
\end{equation}
where $\quad q = (\tau - \tau_j) / (\tau_{j+1} - \tau_j)$.

\paragraph{Cubic.}
The cubic interpolation leverages finite-difference approximations of the slope at the interval points:
\begin{align}
    \phi_j &= \frac{1}{2} \left(\frac{\theta_{j+1} - \theta_j}{
               \tau_{j+1} - \tau_j} +
            \frac{\theta_j - \theta_{j-1}}{
               \tau_j - \tau_{j-1}} \right),\\
    \phi_{j+1} &= \frac{1}{2} \left(\frac{\theta_{j+2} - \theta_{j+1}}{
               \tau_{j+2} - \tau_{j+1}} +
            \frac{\theta_{j+1} - \theta_{j}}{
               \tau_{j+1} - \tau_{j}} \right),
\end{align}
and returns:
\begin{equation}
    a \cdot \theta_j + b \cdot \phi_j + c \cdot \theta_{j+1} + d \cdot \phi_{j+1} \leftarrow s,
\end{equation}
where,
\begin{align}
    a &= 2 q^3 - 3 q^2 + 1,\\
    b &= (q^3 - 2 q^2 + q) \cdot (\tau_{j+1} - \tau_j),\\
    c &= -2 q^3 + 3 q^2,\\
    d &= (q^3 - q^2) \cdot (\tau_{j+1} - \tau_j).
\end{align}

\section{Tasks}
This section provides additional information about the examples provided in Section \ref{results}, including objective formulations and planner settings.

\paragraph{Humanoid objective.} \label{obj_humanoid}
This task comprises $M = 6$ cost terms:
\begin{itemize}
    \item Term $0$:
    \begin{itemize}
        \item[] $\text{r}_0$: lateral center-of-mass position and average lateral feet position alignment 
        \item[] $\text{n}_0$: hyperbolic cosine
        \item[] $w_0$: 100
    \end{itemize}
    \item Term $1$:
    \begin{itemize}
        \item[] $\text{r}_1$: lateral torso position and lateral center-of-mass position alignment
        \item[] $\text{n}_1$: smooth absolute value
        \item[] $w_1$: 1
    \end{itemize}
    \item Term $2$:
    \begin{itemize}
        \item $\text{r}_2$: head height and feet height difference minus target height difference
        \item $\text{n}_2$: smooth absolute value
        \item $w_2$: 100
    \end{itemize}
    \item Term $3$:
    \begin{itemize}
        \item[] $\text{r}_3$: lateral center-of-mass velocity
        \item[] $\text{n}_3$: smooth absolute value
        \item[] $w_3$: 10
    \end{itemize}
    \item Term $4$:
    \begin{itemize}
        \item[] $\text{r}_4$: joint velocity
        \item[] $\text{n}_4$: quadratic
        \item[] $w_4$: 0.01
    \end{itemize}
    \item Term $5$:
    \begin{itemize}
        \item[] $\text{r}_5$: control effort
        \item[] $\text{n}_5$: quadratic
        \item[] $w_5$: 0.025
    \end{itemize}
\end{itemize}

\paragraph{Quadruped objective.} \label{obj_quadruped}
This task comprises $M = 4$ cost terms: 
\begin{itemize}
    \item Term $0$:
    \begin{itemize}
        \item[] $\text{r}_0$: body height and average feet height difference minus target height
        \item[] $\text{n}_0$: quadratic
        \item[] $w_0$: 1
    \end{itemize}
    \item Term $1$:
    \begin{itemize}
        \item[] $\text{r}_1$: body position minus goal position
        \item[] $\text{n}_1$: quadratic
        \item[] $w_1$: 5.0
    \end{itemize}
    \item Term $2$:
    \begin{itemize}
        \item[] $\text{r}_2$: body orientation minus goal orientation
        \item[] $\text{n}_2$: quadratic
        \item[] $w_2$: 1.0
    \end{itemize}
    \item Term $3$:
    \begin{itemize}
        \item[] $\text{r}_3$: control effort
        \item[] $\text{n}_3$: quadratic
        \item[] $w_3$: 0.25
    \end{itemize}
\end{itemize}

\paragraph{Hand objective.} \label{obj_hand}
This task comprises $M = 3$ cost terms: 
\begin{itemize}
    \item Term $0$:
    \begin{itemize}
        \item[] $\text{r}_0$: cube position minus hand palm position
        \item[] $\text{n}_0$: quadratic
        \item[] $w_0$: 20
    \end{itemize}
    \item Term $1$:
    \begin{itemize}
        \item[] $\text{r}_1$: cube orientation minus goal orientation
        \item[] $\text{n}_1$: quadratic
        \item[] $w_1$: 3
    \end{itemize}
    \item Term $2$:
    \begin{itemize}
        \item[] $\text{r}_2$: cube linear velocity
        \item[] $\text{n}_2$: quadratic
        \item[] $w_2$: 10
    \end{itemize}
\end{itemize}

\paragraph{Planner settings.}
We provide the settings used for Predictive Sampling in Table \ref{ps_settings}.

\begin{table}[H] 
    \centering
    \caption{Predictive Sampling settings}
    \begin{tabular}{c c c c c c}
		\toprule
	    \textbf{Task} & $P$ & $N$ & $T$ & $\sigma$ \\
		\toprule
		Humanoid & 3 & 10 & 23 & 0.125 \\
		Quadruped & 3 & 10 & 35 & 0.25 \\
	    Hand & 6 & 10 & 25 & 0.1 \\
		\toprule
    \end{tabular}
    \label{ps_settings}
\end{table}

\section{Predictive Sampling Algorithm}

\begin{algorithm}[H]
	\caption{PredictiveSampling}\label{predictive_sampling_algorithm_numeric}
	\begin{algorithmic}[1]
		\Procedure{OptimizePolicy}{}
		\State task: $f$, $c$, $R$
		\State settings: $T$, $N$, $\sigma$, $s$
		\State initialise: $(\tau, \theta)$
		\State \textbf{for} $k = 0, \dots, \infty$
		\State \indent $(x, \tau) \leftarrow$ get state
		\State \indent $(\bar{\tau}, \theta) \leftarrow \text{resample}$
		\State \indent \textbf{for} $i = 0, \dots, N$ (parallel)
		\State \indent \indent $\tilde{\theta}^{(i)} = \theta + 
        \begin{cases}
            0, \phantom{\mathcal{N}(0, \sigma^2 \cdot I)} i = 0, \\ 
            \mathcal{N}(0, \sigma^2 \cdot I), \phantom{0} \text{else} 
        \end{cases}$
		\State \indent \indent $x^{(i)}_0 = x, \, \tau_0^{(i)} = \tau$,
		\State \indent \indent \textbf{for} $t = 0, \dots, T$
		\State \indent \indent \indent $u^{(i)}_{t} = s(\tau^{(i)}_t; (\bar{\tau}, \tilde{\theta}^{(i)}))$
		\State \indent \indent \indent $c^{(i)}_{t} = c(x^{(i)}_t, u^{(i)}_t; R)$
		\State \indent \indent \indent $(x^{(i)}_{t+1}, \tau^{(i)}_{t+1}) = f(x^{(i)}_t, u^{(i)}_t)$
		\State \indent \indent \textbf{end for}
		\State \indent \textbf{end for} 
		\State \indent $\theta \leftarrow \text{argmin}\left(J(\tilde{\theta}^{(i)})\right)$
		\State \textbf{end for}
		\EndProcedure
		\Procedure{ActionFromPolicy}{}
		\State $(x, \tau) \leftarrow$ get state
        \State $u = s(\tau; (\bar{\tau}, \theta))$
        \State \textbf{return} $u$
		\EndProcedure
	\end{algorithmic}
\end{algorithm}

\section{Compute Resources}
Experiments were performed on a Lenovo ThinkStation P920 with 48GB of memory and an Intel Xeon Gold 6154 72-core CPU. Additional experiments were performed on an Apple MacBook Pro (2021) with 16 GB of memory and an M1 Pro CPU.

	\end{appendices}
	
\end{document}